\documentclass{esannV2}
\usepackage{graphicx,url,hyperref}
\usepackage{xcolor,wrapfig,algorithmic}
\usepackage[scriptsize,tight]{subfigure}
\usepackage{tabularx}
  \newcolumntype{C}[1]{>{\centering\arraybackslash}p{#1}}
  \newcolumntype{R}[1]{>{\raggedleft\arraybackslash}p{#1}}
  \newcolumntype{L}[1]{>{\raggedright\arraybackslash}p{#1}}
\usepackage{booktabs} 
\usepackage[ruled,linesnumbered]{algorithm2e}
\usepackage{amssymb}
\usepackage{amsmath}
\usepackage{mathtools}
\usepackage{xspace}
\usepackage{listings}
\usepackage{floatrow}

\newcommand{\1}{\mbox{$\mathbb{I}$}}

\newfloatcommand{capbtabbox}{figure}[][\FBwidth]
\usepackage{tikz}

\begin{document}

\date{}

\title{Large Language Models for Tuning Evolution Strategies}

%\title{Tuning Evolution Strategies with Large Language Models}

%\title{How to Tune an Evolution Strategy with a \\Large Language Model}

\author{Oliver Kramer
%
% Optional short acknowledgment: remove next line if non-needed
%\thanks{This is an optional funding source acknowledgement.}
%
% DO NOT MODIFY THE FOLLOWING '\vspace' ARGUMENT
\vspace{.3cm}\\
%
% Addresses and institutions (remove "1- " in case of a single institution)
Computational Intelligence Lab\\Department of Computer Science\\Carl-von-Ossietzky University of Oldenburg \\
26111 Oldenburg, Germany\\
Email: \url{oliver.kramer@uni-oldenburg.de}\\
%\url{oliver.kramer@uni-oldenburg.de}
%
% Remove the next three lines in case of a single institution
}

\maketitle
\thispagestyle{empty}

\begin{abstract}
Large Language Models (LLMs) exhibit world knowledge and inference capabilities, making them powerful tools for various applications. This paper proposes a feedback loop mechanism that leverages these capabilities to tune Evolution Strategies (ES) parameters effectively. The mechanism involves a structured process of providing programming instructions, executing the corresponding code, and conducting thorough analysis. This process is specifically designed for the optimization of ES parameters. The method operates through an iterative cycle, ensuring continuous refinement of the ES parameters. First, LLMs process the instructions to generate or modify the code. The code is then executed, and the results are meticulously logged. Subsequent analysis of these results provides insights that drive further improvements. An experiment on tuning the learning rates of ES using the LLaMA3 model demonstrate the feasibility of this approach. This research illustrates how LLMs can be harnessed to improve ES algorithms' performance and suggests broader applications for similar feedback loop mechanisms in various domains.
\end{abstract}

\section{Introduction}

Large Language Models (LLMs) are advanced models designed for a wide array of natural language processing (NLP) tasks. With a transformer-based architecture containing billions of parameters, LLMs deliver comprehensive NLP capabilities, offering nuanced text understanding, generation, and inference across a broad spectrum of domains.
LLM can be used to autonomously generate and execute programming and tunings tasks, advancing the potential for self-improving algorithms. 

This paper proposes an iterative approach using LLM to tune the parameters of ES \cite{schwefel}. By automating the process of refining and tuning algorithms, LLMs facilitate the development of more sophisticated systems capable of autonomous learning, adaptation, and execution. 

The paper is structured as follows. Section \ref{sec:related} summarizes related work. Section \ref{sec:llmes} introduces the iterative LLM-based parameter tuning approach for ES. An experimental analysis is presented in Section \ref{sec:exp}. Conclusions are drawn in Section \ref{sec:cons}.

\section{Related Work}
\label{sec:related}

Tuning evolutionary algorithm parameters has a rich history, with recent advancements leveraging LLMs for hyperparameter optimization and algorithm adaptation. Zhang et al. \cite{Zhang2023a} demonstrated LLMs' efficacy in hyperparameter optimization by treating model code as a hyperparameter. Similarly, Zheng et al.~\cite{Zheng} showcased GPT-4's utility in Neural Architecture Search (NAS). DeepMind's OptFormer \cite{optformer} introduced a transformer-based framework for hyperparameter optimization. LLM-driven Evolutionary Algorithms (LMEA) \cite{Liu2023} demonstrated competitive performance on combinatorial problems like the Traveling Salesman Problem. AutoML-GPT \cite{automlgpt} automated the training pipeline using LLMs, achieving strong results across various domains. Guo et al. \cite{guo} directly applied LLMs to optimization tasks, noting their effectiveness on small datasets. The Language-Model-Based Evolutionary Optimizer (LEO) \cite{Brahmachary2023} applied LLMs to numerical optimization tasks but requires careful handling to mitigate hallucination risks. Despite these advancements, no explicit approach has focused on tuning ES parameters. This paper addresses this gap by proposing a feedback loop mechanism involving instructions for programming, execution, and analysis of code, specifically for tuning ES parameters.

\section{LLM Tuning Evolution Strategies}
\label{sec:llmes}

\begin{figure}[h!]
\centering
\includegraphics[width=\textwidth]{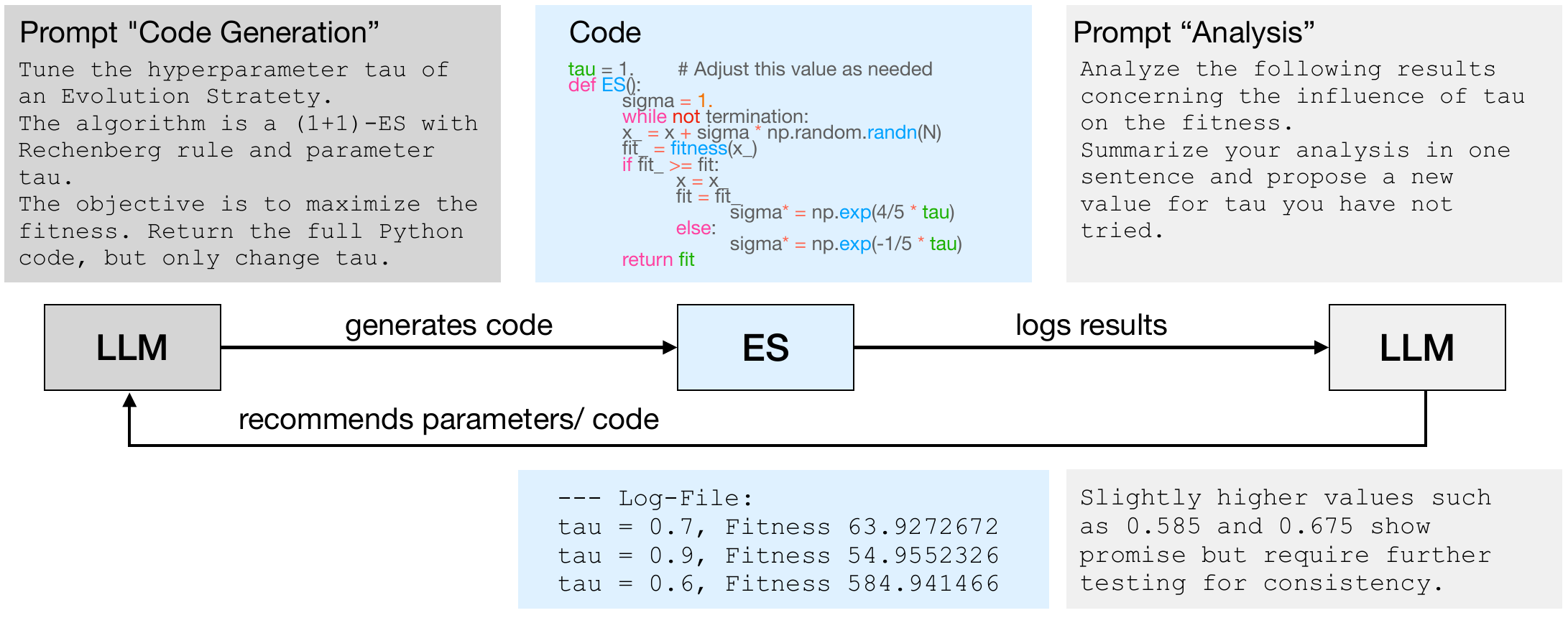}
\caption{\label{fig:llmes}Overview of LLM and ES interaction with exemplary prompts and code.} 
\end{figure}

We propose a system that autonomously generates and executes programming tasks, advancing the potential for self-improving algorithms. It interprets high-level instructions, in particular for tuning ES parameters and translates them into actionable code, runs it, and evaluates its performance, see Figure \ref{fig:llmes} for an overview. 

In general, the system is designed to handle natural language prompts or predefined instructions, translating them into functional code for execution. Leveraging an NLP engine, the input is converted into programming scripts which are then run with exceptions handling. Throughout execution, key metrics like fitness function evaluations and solution quality w.r.t. tunable parameters are logged, allowing for post-execution analysis to evaluate performance. Based on this analysis, the system iteratively refines its parameters and the corresponding code, facilitating ongoing improvements. The concrete implementation is depicted in the following.

\subsection{Evolution Strategies}
\label{sec:es}

Evolution Strategies (ES) are optimization algorithms inspired by natural evolution, evolving a population of candidate solutions to tackle complex optimization problems iteratively. Among these strategies, the (1+1)-ES with Rechenberg rule is a fundamental variant. This method begins with an initial solution vector \(\mathbf{x}\) and a mutation scaling factor \(\sigma\). Each iteration involves generating a new candidate solution \(\mathbf{x}'\) by adding Gaussian noise to \(\mathbf{x}\), with the noise scaled by \(\sigma\). The algorithm accepts the new solution \(\mathbf{x}'\) if it improves upon the current solution \(\mathbf{x}\) according to the fitness function \(f\). When an improvement is made, \(\sigma\) is adjusted exponentially using the 1/5th rule, which is designed to maintain a balance between exploration and exploitation. The mutation parameter \(\tau\) plays a crucial role in this adaptive mechanism, ensuring that the search process can efficiently navigate the solution space. The adaptation of the step size \(\sigma\) follows the formula:
\begin{equation}
\sigma := \sigma \cdot \exp(\tau \cdot (\1_{f(\mathbf{x}') \leq f(\mathbf{x})} - 1/5 ))
\end{equation}
In this equation, \(\1_{f(\mathbf{x}') \leq f(\mathbf{x})}\) yields 1 if \(f(\mathbf{x}') \leq f(\mathbf{x})\), and 0 otherwise. This mechanism ensures that the step size increases when improvements are consistently found, and decreases otherwise, allowing the algorithm to dynamically adjust its search strategy until a termination condition is met.

\subsection{LLM Instructions}

The LLM is instructed to explore new values for $\tau$ that haven't been previously tested, based on its analysis. The modified \texttt{Python} code should reflect this new value for~$\tau$, facilitating an iterative tuning process that optimizes the algorithm's performance for the given context. A prompt should provide clear context and outline the algorithm and parameters targeted for tuning. Furthermore, it's advisable to recommend experimenting with novel parameter values while refraining from modifying the source code elsewhere. The prompt guiding the LLM in the experimental section is the following:
%\begin{small}
\begin{lstlisting}
Tune the hyperparameter tau of an Evolution Stratety.
The algorithm is a (1+1)-ES with Rechenberg rule and parameter tau.
The objective is to maximize the fitness.
Return the full Python code, but only change tau.
\end{lstlisting}
%\end{small}

\subsection{Execution Process}

\texttt{Python}  code is executed, with artifacts like "code" or "python" removed to ensure a clean execution environment. To mitigate program crashes, exception handling via `\texttt{try \ldots except}` blocks is employed, effectively capturing potential errors and managing them gracefully.

\subsection{Log File}

The log file of results is basis of the feedback analysis process. It is initially empty, then filled with results that capture the most important aspects. The following example logs the settings an ES with parameter $\tau$ and the resulting fitness of 10 runs:
\begin{lstlisting}
tau = 0.7, Fitness: 0.1162058339177609
tau = 0.95, Fitness: 66.05538351053897
\end{lstlisting}
Statistical properties may be mentioned here as well if they might improve the following analysis process.

\subsection{Analysis}

The analysis and feedback mechanism harnesses the capabilities of an LLM to derive insights from the log file. When coupled with the contextual guidance provided in the instruction prompts, this approach yields valuable parameter recommendations. An example for a useful prompt for the log file analysis and proposal of new parameters is:
\begin{lstlisting}
Analyze the following results concerning the influence of tau on the fitness.
Summarize your analysis in one sentence and propose a new value for tau you have not tried.
\end{lstlisting}
This approach enables automated, succinct interpretation of results. An exemplary result interpretion is shown in the experimental section.

\section{Experimental Study}
\label{sec:exp}

In the following, we present an exemplary experiment to demonstrate the feasibility of our approach, utilizing the LLaMA3 model for updating the \texttt{Python} code of an ES and for analyzing the experimental results. The experiment involves a (1+1)-ES with Rechenberg's rule applied to a 5-dimensional Sphere function. This scenario focuses on tuning the parameter \(\tau\) to observe its impact on the algorithm's performance.

\begin{figure}[h!]
\centering
\includegraphics[width=0.6\textwidth]{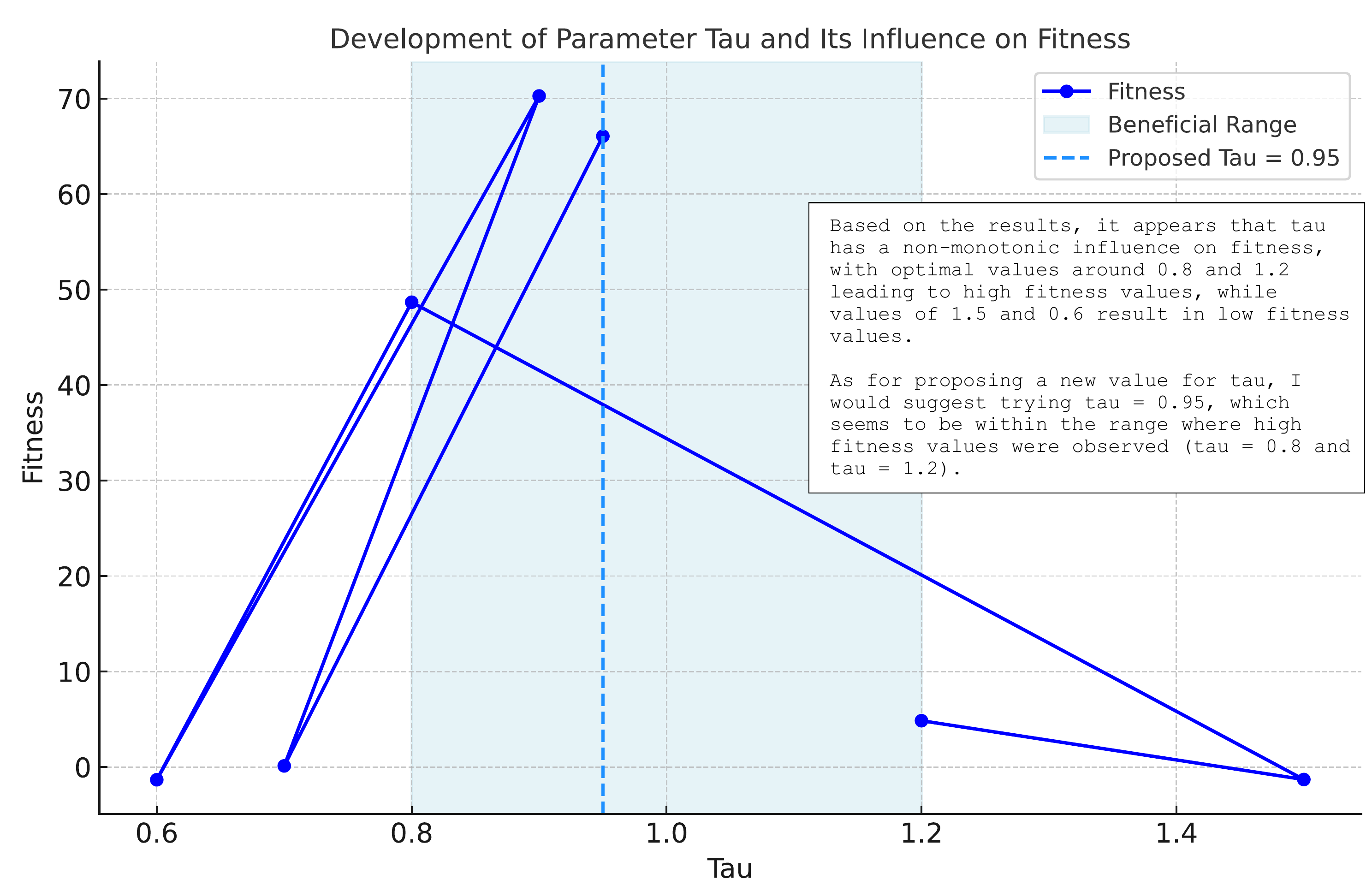}
\caption{\label{fig:exp}The plot illustrates that the LLM finds the optimal range for the parameter \(\tau\) in the scenary of the (1+1)-ES on a 5-dimensional Sphere function (here maximizing $- \log f(\mathbf{x})$) with a proposed optimal value of \(\tau = 0.95\).} 
\end{figure}

The results are summarized as follows, see Figure \ref{fig:exp} presenting a typical run. The LLM experimented with various values of \(\tau\) and measured the corresponding fitness outcomes. The plot depicts the relationship between the parameter \(\tau\) and the fitness values achieved during the optimization process running the (1+1)-ES on the 5-dimensional Sphere function for 1000 generations, with the mean of 10 runs analyzed. The x-axis represents various \(\tau\) values, ranging from 0.6 to 1.5, while the y-axis shows the corresponding fitness values as $- \log f(\mathbf{x})$.

The figure illustrates an analytical step where the LLM concludes that \(\tau\) values between 0.8 and 1.2 are associated with higher fitness, indicating this range is beneficial for the optimization process. In contrast, higher values for \(\tau\) result in significantly lower fitness, suggesting a decline in performance. The proposed new value of \(\tau = 0.95\) falls within the beneficial range, indicating it may offer an optimal balance between exploration and exploitation.

\section{Conclusions}
\label{sec:cons}

In conclusion, our experimental study demonstrates the feasibility and effectiveness of using the LLaMA3 model for updating \texttt{Python} code within optimization scenarios. Specifically, the (1+1)-ES with Rechenberg's rule applied to a 5-dimensional Sphere function shows that the choice of the parameter \(\tau\) significantly influences the algorithm's performance and LLaMA3 is able to find good settings. These findings emphasize the importance of parameter tuning in optimization strategies and highlight the ability of LLMs to tune hyperparameters.

Moreover, this approach opens new avenues for future research in autonomous optimization. The techniques outlined can be applied to a variety of domains, including intricate programming tasks. This broad applicability underscores the potential of LLMs to revolutionize how algorithms are developed and refined autonomously, making significant contributions to the field of optimization.

\appendix
\section{Benchmark Function}

The Sphere function is 
\begin{equation}
f(\mathbf{x}) = \sum_{i=1}^N x_i^2,
\end{equation}
where \(\mathbf{x} \in \mathbb{R}^N\) with a global optimum at \(\mathbf{x}^* = \mathbf{0}\) with \(f(\mathbf{x}^*) = 0$.

\bibliographystyle{abbrv}

\end{document}